# Causes of Ineradicable Spurious Predictions in Qualitative Simulation

**Özgür Yılmaz**  YILMOZGU@BOUN.EDU.TR
**A. C. Cem Say**  SAY@BOUN.EDU.TR
*Department of Computer Engineering*
*Boğaziçi University*
*Bebek 34342 İstanbul, Turkey*

**Abstract**

It was recently proved that a sound and complete qualitative simulator does not exist, that is, as long as the input-output vocabulary of the state-of-the-art QSIM algorithm is used, there will always be input models which cause any simulator with a coverage guarantee to make spurious predictions in its output. In this paper, we examine whether a meaningfully expressive restriction of this vocabulary is possible so that one can build a simulator with both the soundness and completeness properties. We prove several negative results: All sound qualitative simulators, employing subsets of the QSIM representation which retain the operating region transition feature, and support at least the addition and constancy constraints, are shown to be inherently incomplete. Even when the simulations are restricted to run in a single operating region, a constraint vocabulary containing just the addition, constancy, derivative, and multiplication relations makes the construction of sound and complete qualitative simulators impossible.

## 1. Introduction

It was recently proved (Say & Akın, 2003) that a sound and complete qualitative simulator does not exist, that is, as long as the input-output vocabulary of the state-of-the-art QSIM algorithm (Kuipers, 1994) is used, there will always be input models which cause any simulator with a coverage guarantee to make spurious predictions in its output. In this paper, we examine whether a meaningfully expressive restriction of this vocabulary is possible so that one can build a simulator which will always output all and only the consistent solutions of its input model. We prove several negative results: All sound qualitative simulators, employing subsets of the QSIM representation which retain the operating region transition feature, and support at least the addition and constancy constraints, are shown to be inherently incomplete. The problem persists when all variables are forced to change continuously during region transitions if a slightly larger set of constraint types is allowed. Even when the simulations are restricted to run in a single operating region, a constraint vocabulary containing just the addition, constancy, derivative, and multiplication relations makes the construction of sound and complete qualitative simulators impossible. Our findings may be helpful for researchers interested in constructing qualitative simulators with improved theoretical coverage guarantees using weaker representations.

## 2. Background

We start with a brief overview of qualitative simulation, concentrating on the representations used in the input-output vocabularies of qualitative simulators. Subsection 2.2 summarizes previous work on the two theoretical properties of qualitative simulators that interest us. Subsection 2.3 is a short "requirements specification" for a hypothetical sound and complete qualitative simulator.

## 2.1 Qualitative Simulation

In many domains, scientists and engineers have only an incomplete amount of information about the model governing the dynamic system under consideration, which renders formulating an exact ordinary differential equation (ODE) impossible. Incompletely specified differential equations may also appear in contexts where the aim is to find collective proofs for behavioral properties of an infinite set of systems sharing most, but not all, of the structure of the ODEs describing them. To proceed with the reasoning task in such cases, mathematical tools embodying methods making the most use of the available information to obtain a (hopefully small) set of possible solutions matching the model are needed. *Qualitative reasoning* (QR) researchers develop AI programs which use "weak" representations (like intervals rather than point values for quantities, and general shape descriptions rather than exact formulae for functional relationships) in their vocabularies to perform various reasoning tasks about systems with incomplete specifications. In the following, we use the notation and terminology of QSIM (Kuipers, 1994), which is a state-of-the art qualitative simulation methodology, although it should be noted that the incompleteness results that we will be proving are valid for all reasoners whose input-output vocabularies are rich enough to support the representational techniques that will be used in our proofs.

A qualitative simulator takes as input a *qualitative differential equation* model of a system in terms of constraints representing relations between the system's variables. In addition to this model, the qualitative values of the variables at the time point from which the simulation should start are also given. The algorithm produces a list of the possible future behaviors that may be exhibited by systems whose ordinary differential equations match the input model.

The *variables* of a system modeled in QSIM are continuously differentiable functions of time. The limits of each variable and their first derivatives exist as they approach the endpoints of their domains. Each variable has a *quantity space*; a totally ordered collection of symbols (*landmarks*) representing important values that it can take. Zero is a standard landmark common to all variables. Quantity spaces are allowed to have the landmarks -∞ and ∞ at their ends, so functional relationships with asymptotic shapes can be explicitly represented. When appropriate, a quantity space can be declared to span only a proper subset of the extended reals; for instance, it makes sense to bound the quantity space for a variable which is certainly nonnegative (like pressure) with 0 at the left. If necessary, the user can specify one or both bounds of a quantity space to be unreachable; for example, ∞ will be an unreachable value for all variables in all the models to be discussed in this paper. The (reachable) points and intervals in its quantity space make up the set of possible *qualitative magnitudes* of a variable. The *qualitative direction* of a variable is defined to be the sign of its derivative; therefore its possible values are: *inc* (+), *dec* (−) and *std* (0). A variable's *qualitative value* is the pair consisting of its qualitative magnitude and qualitative direction. The collection of the qualitative values of its variables makes up the *state* of a system.

The "laws" according to which the system operates are represented by *constraints* describing time-independent relations between the variables. At each step of the simulation, QSIM uses a set of transition rules to implicitly generate all possible "next" values of the variables. The combinations of these values are filtered so that only those which constitute complete states, in which every constraint is still satisfied by the new values of its variables, remain.

There are seven "basic" types of constraints in QSIM. (See Table 1.) Each type of constraint imposes a different kind of relation on its arguments. For example, if we have the constraint $A(t) = -B(t)$, any combination of variable values in which variables $A$ and $B$ have the same (nonzero) sign in their magnitudes or directions will be filtered out. Sometimes, additional knowledge about the constraints allows further filtering. In the above example, if we know that $A$ and $B$ had the landmark values $a_1$ and $b_1$ at the same moment at some time in the past, we can eliminate all value combinations in which *A* and *B* have magnitudes both less (or both greater) than these landmarks. $a_1$ and $b_1$ are called *corresponding values* of that constraint, and the equation $a_1 = -b_1$ is a *correspondence.* Each constraint in a model (except those of the *derivative* type) can have such correspondence equations. A "sign algebra" (Kuipers, 1994) is employed to implement the arithmetic relations using qualitative magnitudes. Note that, since each $M^+$ ($M^-$) relationship corresponds to an infinite number of possible "quantitative" functions having the monotonicity property, a single QSIM model containing such constraints can correspond to infinitely many ODEs.

| CONSTRAINT NAME | NOTATION | EXPLANATION |
|---|---|---|
| add | $X(t) + Y(t) = Z(t)$ | |
| constant | $X(t)$ = a landmark | $\frac{d}{dt} X(t) = 0$ |
| derivative | $d/dt(X,Y)$ | $\frac{d}{dt} X(t) = Y(t)$ |
| $M^+$ | $X(t) = f(Y(t))$, $f \in M^+$ | $\exists f$ such that $X(t) = f(Y(t))$, where $f' > 0$ over $f$'s domain |
| $M^-$ | $X(t) = f(Y(t))$, $f \in M^-$ | $\exists f$ such that $X(t) = f(Y(t))$, where $f' < 0$ over $f$'s domain |
| minus | $X(t) = -Y(t)$ | |
| mult | $X(t) \times Y(t) = Z(t)$ | |

Table 1: The Qualitative Constraint Types

The QSIM input vocabulary enables the user to describe more complicated models in terms of several different constraint sets representing different *operating regions* of the system under consideration. The user specifies the boundaries of the applicability ranges of the operating regions in terms of conditions which indicate that the simulator should effect a transition to another operating region when they are obtained.

For each operating region from which such a transition can occur, one has to specify the following for each possible transition:
- Boolean expressions composed of primitives of the form "*VariableName=QualitativeValue*", which will trigger this transition when they are satisfied,
- The name of the target operating region,
- The names of variables which will inherit their qualitative magnitudes and/or directions in the first state after the transition from the last state before the transition,
- Value assignments for any variables which will have explicitly specified values in the first state after the transition.

When provided with a qualitative system model, the name of the initial operating region, and a description of the qualitative values of all variables in the initial state, QSIM starts simulation, and generates a tree of system states to represent the solutions of the qualitative differential equation composed of the constraints in its input. The root of this tree is the input initial state with the time-point label $t_0$, representing the numerical value of the initial instant. Every path from the root to a leaf is a predicted *behavior* of the system. Being in the qualitative format, each such behavior usually corresponds to an infinite set of trajectories sharing the same qualitative structure in phase space. Time-point and interval states appear alternately in behaviors as long as the same operating region is valid. Operating region transitions are reflected in behaviors as two time-point states following each other.

### 2.2 Related Work on Soundness and Incompleteness

A very important property of qualitative simulators is their "coverage guarantee": A qualitative simulation algorithm is *sound* if it is guaranteed that, for any ODE and initial state that matches the simulator's input, there will be a behavior in its output which matches the ODE's solution. Kuipers (1986) proved that there exists a qualitative simulator (namely, QSIM) that has the soundness property. This guarantee makes qualitative simulation a valuable design and diagnosis method (Kuipers, 1994): In design, if the set of simulation predictions of our model does not contain a catastrophic failure, this is a *proof* that our modeled system will not exhibit that failure (Shults & Kuipers, 1997). In diagnosis, if none of the behaviors in the simulation output of a model is being exhibited by a particular system, we can be 100% sure that the actual system is *not* governed by that model.

Another property that one would wish one's qualitative simulator to possess is *completeness*; that is, a guarantee that every behavior in its output corresponds to the solution of at least one ODE matching its input. In the early days of QR research, it was conjectured (de Kleer & Brown, 1984) that qualitative simulators employing local constraint satisfaction methods (Weld & de Kleer, 1990) were complete. However, in the same paper which contained the guaranteed coverage theorem, Kuipers (1986) also showed that the version of QSIM described there, and, indeed, all qualitative simulators of the day, were *incomplete*, by demonstrating that the simulation of a frictionless mass-spring oscillator predicts unrealizable (*spurious*) behaviors, where the amplitude decreases in some periods and increases in others. The lack of a guarantee that all the predicted behaviors are real has a negative impact on potential applications: In design, if the set of simulation predictions of our model *does* contain a catastrophic failure, this does not necessarily point to an error in our mechanism; maybe the prediction in question was just a spurious behavior. A similar problem occurs in diagnosis applications.

Several other types of spurious qualitative simulation predictions were discovered in the following years: Struss (1990) pointed out that, whenever a variable appeared more than once in an arithmetic constraint, spurious states could pass the filter. For instance, the filters of the *add* constraint are unable to delete states involving nonzero values for the variable $Z$ in the equation $A(t) + Z(t) = A(t)$ when $A$ is nonzero. Clearly, any sound and complete qualitative simulator would have to possess the algebraic manipulation capabilities that enable us to conclude that $Z=0$ in this case. Say and Kuru (1993) discovered a class of spurious predictions caused by a rigidity in the internal representation of correspondences, and an unnecessarily weak implementation of subtraction. Say (1998) showed that some other spurious behaviors are due to a lack of explicit enforcement of l'Hôpital's rule in the original algorithm. Yet another family of inconsistent predictions was found out to be caused by weaknesses in the methods used to distinguish finite and infinite time intervals in the behaviors (Say, 2001, also see Missier, 1991). Könik and Say (2003) proved that some model and behavior descriptions could "encode" information about the relative (finite) lengths of the intervals that they contain, and failure to check the overall consistency of these pieces of information yields another class of spurious predictions. Finally, Say (2003) showed that a similar encoding could occur about the exact numerical values of some landmarks, so that a sound and complete qualitative simulator would have to support a capability of comparing the magnitudes of any two elements of a very rich subset of the real numbers to avoid a particular set of spurious predictions.

Interestingly, all these discoveries were actually good news for the users of qualitative simulators: In order to be able to say that a particular predicted behavior is spurious, and therefore suitable for elimination from the simulator output without forsaking the soundness property, one first proves that that behavior is mathematically inconsistent with the simulated model and starting state. For instance, the aforementioned spurious oscillations of the frictionless mass-spring system can be shown to violate a conservation constraint that follows directly from the structure of the input equations. But this proof can itself be seen as the specification of a new filter routine which would eliminate exactly the set of behaviors that violate the "law" that it establishes. The kinetic energy constraint (Fouché & Kuipers, 1992) is a filter which has been developed in this fashion to eliminate the class of spurious predictions exemplified by the ones about the mass-spring system (Kuipers 1994). So all the spurious prediction classes mentioned in the previous paragraph had, in fact, been discovered simultaneously with their "cures."

The question of whether there exists a sound and complete qualitative simulator was finally settled by Say and Akın (2003). They proved that, for any sound qualitative simulator using the input-output representation and task specification of the QSIM methodology, there exist input models and initial states whose simulation output will contain spurious predictions. (Note that this does not just say that the present QSIM algorithm can not be augmented by more filters to make it both sound and complete; it refutes the existence of any program whatsoever that can perform this job.)

The proof by Say and Akın (2003) shows that a sound and complete qualitative simulator employing the vocabulary mentioned above, if it existed, could be used to solve any given instance of Hilbert's Tenth Problem, which is famously undecidable (Matiyasevich, 1993). The procedure involves building a QSIM model representing the given problem, simulating it several times starting from carefully constructed initial states representing candidate solutions, and examining the output to read out the solution. The model is set up to contain an inconsistency if and only if the answer to the considered problem is "no," so the very existence of one or more behaviors in the output means "yes." Since it is impossible to make this decision correctly in the general case, it follows that there would be input models giving rise to behavior predictions whose consistency status can not be determined by the simulator, whose best course of action would be to include them in the output, to keep the soundness guarantee intact. In the cases where the correct answer is "no," this would result in the prediction of spurious behaviors. Note that these are *ineradicable* spurious predictions, unlike the ones discussed earlier.

It is important to note that this proof does not necessarily mean that all hope of constructing a sound and complete qualitative simulator is lost. One may try to "weaken" the input-output representation so that it no longer possesses the problematic power which enables one to unambiguously encode instances of Hilbert's Tenth Problem into a QSIM model. (Of course, this weakening must be kept at the minimum possible level for the resulting program to be a useful reasoner; for instance, removing the program's ability to distinguish between negative and nonnegative numbers would possibly yield a sound and complete simulator, but the output of that program would just state that "everything is possible" and this is not what we want from these methods.) This is why one should examine the incompleteness proof (Say & Akın, 2003) to see exactly which features of the QSIM representation are used in the construction of the reduction; any future qualitative simulator supporting the same vocabulary subset would be incorporating the same problem from the start.

Here is a listing of the QSIM representational items used in that proof: Only the $M^+$, *derivative*, *mult*, and *constant* constraint types are utilized. (Note the absence of the *add* constraint, which can be "implemented" using the others, in this list.) Qualitative interval magnitudes like (0, ∞), with what one might call "infinite uncertainty" about the actual value of the represented number, are used for initializing several variables, and form an essential part of the argument. QSIM's ability to explicitly represent infinite limits is utilized for equating a landmark to the number π, by stating that it is twice the limit of the function *arctan x* as *x* nears infinity. Finally, the operating region transition feature is used heavily, since it is thanks to this characteristic that the sine function between two dependent variables can be represented in the qualitative vocabulary.

In Section 3, we examine several different ways of weakening the QSIM vocabulary to try to understand which combinations of these features are responsible for the problem of ineradicable spurious predictions.

### 2.3 Desiderata for a Sound and Complete Qualitative Simulator

It is important at this point to clarify exactly what one would expect from a hypothetical sound and complete qualitative simulator. If the input model yields a finite behavior tree of genuine solutions, it is obvious that the program is supposed to print the descriptions of the behaviors forming the branches of this tree, and nothing else, in finite time. If the input model and initial state are inconsistent, i.e., the "correct" output is the empty tree, the program should report this inconsistency in finite time.

Finally, if the input yields a behavior tree with infinitely many branches, the program is supposed to run forever, adding a new state to its output every once in a while. More formally, for every positive $i$, there has to be an integer $s$ such that the program will have printed out the "first" $i$ states of the behavior tree (according to some ordering where the root, i.e. the initial state, is state number 1, and no descendants of any particular state are printed before that state itself) at the end of the $s^{th}$ step in its execution. Note that these requirements mean that a sound and complete simulator would have to be able to decide whether the initial system state description given to it is consistent with the input model or not within finite time. This necessity is used in the proofs of incompleteness in Section 3.

## 3. New Incompleteness Results for Qualitative Simulators

In this section, we examine two different ways of restricting the qualitative vocabulary in the hope of obtaining a representation which allows the construction of sound and complete simulators. Subsection 3.1 considers the usage of several reduced sets of qualitative constraint types, while retaining the operating region transition feature. We also examine a possible restriction on the way variable values are handled during operating region transitions. Subsection 3.2 is an investigation of the capabilities of qualitative simulators which are restricted to input models with a single operating region. Both variants are shown to exhibit the problem of ineradicable spurious predictions when the soundness guarantee is present.

### 3.1 Reduced Constraint Sets

The results in this subsection are based on the undecidability properties of abstract computational devices called unlimited register machines (URMs). We first present a brief introduction to URMs, and then proceed with our proofs.

### 3.1.1 UNLIMITED REGISTER MACHINES

The easiest way of thinking about a URM is to see it as a computer with infinite memory which supports a particularly simple programming language. A URM (Cutland, 1980) program P consists of a finite sequence of instructions $I_1, I_2, ..., I_{|P|}$. The instructions refer to the machine's registers $R_i$, each of which can store an arbitrarily big natural number. We use the notation $r_1, r_2, r_3, ...$ for the register contents.

For our purposes, it will be sufficient to consider three types of URM instructions:

*succ(n)*: Increment the content of register *n* by one.
$$R_n \leftarrow r_n + 1$$
*zero(n)*: Set the content of register *n* to zero.
$$R_n \leftarrow 0$$
*jump(m, n, q)*: Compare registers *m* and *n*. If they are equal, continue with instruction *q*.
$$\text{If } r_m = r_n \text{ then jump to } I_q$$

A URM program starts execution with the first instruction. If the current instruction is not a *jump* whose equality condition is satisfied, it is followed by the next instruction in the list. The program ends if it attempts to continue beyond the last instruction, or if a *jump* to a nonexistent address is attempted. We assume without loss of generality that all such *jump*s are to address $I_{|P|+1}$.

If $P = I_1, ..., I_{|P|}$ is a URM program, it computes a function $P^{(k)} : N^k \rightarrow N$. $P^{(k)}(a_1, ..., a_k)$ is computed as follows:
- <u>Initialization</u>: Store $a_1, ..., a_k$ in registers $R_1, ..., R_k$, respectively, and set all other registers referenced in the program to 0.
- <u>Iteration</u>: Starting with $I_1$, execute the instructions in the order described above.
- <u>Output</u>: If the program ends, then the computed value of the function is the number $r_1$ contained in register $R_1$. If the program never stops, then $P^{(k)}(a_1, ..., a_k)$ is undefined.

Table 2 contains an example URM program which computes the function $f(x, y) = x + y$. Note that the function is from $N^2$ to $N$, where the input values *x* and *y* are stored in registers $R_1$ and $R_2$, and the output of the function is expected to be stored in $R_1$ at the end of the program.

| | |
|---|---|
| $I_1$: | *zero*(3) |
| $I_2$: | *jump*(2, 3, 6) |
| $I_3$: | *succ*(1) |
| $I_4$: | *succ*(3) |
| $I_5$: | *jump*(1, 1, 2) |

Table 2: URM Program Computing $f(x, y) = x + y$

The program first sets $R_3$ to zero. It then checks to see if $R_2 = R_3$ (in the case that $y = 0$). Otherwise, it increments both $R_1$ and $R_3$. This continues until *x* has been incremented *y* times, and the value in $R_1$ is returned.

The URM model of computation is equivalent to the numerous alternative models such as the Turing machine model, the Gödel-Kleene partial recursive functions model and Church's lambda calculus, (Cutland, 1980; Shepherdson & Sturgis, 1963) in the sense that the set of functions computable by URMs is identical to the set of the functions that can be computed by any of the other models. This means that a device which can simulate any given URM is as powerful as a Turing machine, since it can simulate any given Turing machine. In our proofs of the incompleteness of the QSIM vocabulary with reduced constraint sets, we will make use of the fact that the halting problem for URMs is undecidable (Cutland, 1980).

### 3.1.2 Sound and Complete QSIM with Reduced Constraint Sets Solves the Halting Problem

All the incompleteness results about new subsets of the QSIM vocabulary that are presented in this subsection are based on the following theorem, which shows that QSIM can simulate any URM, and thereby has Turing-equivalent computational power.

**Theorem 1:** The execution of any URM program P with |P| instructions on any given input can be represented by the simulation of a QSIM model with |P|+2 operating regions.

**Proof:** The proof will be by construction. Suppose we are given a URM program P with instructions $I_1, ..., I_{|P|}$. Let $R_1, ..., R_N$ be the registers mentioned in the instructions of P.

For each of the $R_i$, we define a QSIM variable $NR_i$ which will represent it, and, in case $R_i$ has a nonzero initial value $a_i$, a set of auxiliary QSIM variables for representing $a_i$. Table 3 describes the idea behind this representation. We have four more variables named *U*, *W*, *Z*, and *B*. *U* represents a "clock" which rises from 0 to 1 in every computational step. *W* is the derivative of *U*. *Z* is constant at zero in every operating region. *B* is an additional auxiliary variable.

| CONSTRAINTS | INITIAL VALUES | CONCLUSIONS |
|---|---|---|
| $ONE(t) \times ONE(t) = ONE(t)$ | $ONE(t_0) = one \in (0, \infty)$ | *ONE* is constant at 1 |
| $ONE(t) + ONE(t) = TWO(t)$ | | *TWO* is constant at 2 |
| $ONE(t) + TWO(t) = THREE(t)$ | | *THREE* is constant at 3 |

Table 3: QSIM Model Fragment Demonstrating the Representability of Exact Integer Values

Our QSIM model will have |P|+2 operating regions: Each instruction $I_i$ of P will have a corresponding operating region named $OpReg_i$. The two remaining regions are $OpReg_0$, corresponding to the "initialization" stage of P, and $OpReg_{|P|+1}$, corresponding to its end.

The specification of each operating region must contain the constraints that are valid in that region, the Boolean conditions (if any) composed of primitives of the form *Variable = <qualitative magnitude, qualitative direction>* which would trigger transitions to other operating regions when they are obtained, and lists that detail which variables inherit their previous magnitudes and/or directions after such a transition, and which of them are initialized to new values during that switch. Tables 4-9 describe how to prepare these items for the operating regions in our target model, based on the program P. There are five different operating region templates (or "types") used in the construction; one for each URM instruction type, one for $OpReg_0$, and one for $OpReg_{|P|+1}$.

The model of $OpReg_0$ is depicted in Table 4. This is where our simulation of P will start. All the $NR_i$ variables are equated to their proper initial values specified by the "user" of P: The ones initialized to zero are handled by "constant at 0" constraints. The ones with positive initial values are specified to be constant at those values using *add* constraints to link them to the "number" variables exemplified in Table 3; for instance, if $NR_2$ is to be initialized to three, we have the constraint $THREE(t) + Z(t) = NR_2(t)$. The *add* constraint between *B*, *U*, and *ONE* serves to express the fact that the landmark named *one* in *U*'s quantity space is also equal to 1. (Note that all the *add* constraints mentioned in this paragraph exist only in $OpReg_0$, since they would disrupt the intended behavior in the other operating regions.)

As seen in Tables 4-8, exactly which variables keep their values during a transition depends on the type of the target operating region. Regions corresponding to instructions of the type *zero*(*n*) should not inherit the value of $R_n$ from their predecessors, since they involve the replacement of that value by zero anyway. All other types of regions, including the *succ*(*n*) type, inherit all the register contents from their predecessors. (Although the value of $R_n$ *does* change in a *succ* instruction, the new value depends on the old one, unlike the case of *zero*(*n*). The corresponding QSIM variable $NR_n$ increases continuously during the simulation of a region of type *succ*(*n*), and a new region transition occurs exactly at the moment when it has increased by one unit.)

---

**Operating Region:** $OpReg_0$

**{Type:** *Initialization*}

**Constraint Set:** $d/dt(U, W)$

$B(t) + U(t) = ONE(t)$ with correspondence "0 + *one* = *one*"

All the required "number representation" constraints (see Table 3)

*add* constraints linking the $NR_i$ to the relevant "number" variables (see text)

All variables except *U* and *B* are constant.

**Possible Transition:**

    **Trigger:** ( $U$ = <*one, inc*> )

    **New operating region:** $OpReg_1$

    **Variables inheriting magnitudes or directions:** See Table 5, indexed by the type of $OpReg_1$

    **Variables with new asserted values:** $U \leftarrow$ <0, *inc*>

---

Table 4: Model of the Operating Region $OpReg_0$, Corresponding to the Initialization of the URM

| TYPE OF TARGET REGION | VARIABLES INHERITING QUALITATIVE MAGNITUDES | VARIABLES INHERITING QUALITATIVE DIRECTIONS |
|---|---|---|
| *jump*(*m, n, q*) | All variables except *U* and *B* | All variables except *U*, *B*, and all the $NR_i$ |
| *succ*(*n*) | All variables except *U* and *B* | All variables except *U*, *B*, and all the $NR_i$ |
| *zero*(*n*) | All variables except *U* and $NR_n$ | All variables except *U*, *B*, and all the $NR_i$ |
| End | All variables except *U* | All variables except *U* and all the $NR_i$ |

Table 5: Variables which should Inherit Magnitudes and/or Directions According to the Type of the Target Operating Region

YILMAZ & SAYThe simulation of the given URM program proceeds as follows: As described in the previous subsection, the URM starts with an initial configuration, where the registers $R_1, ..., R_k$ store the nonnegative integers $a_1, ..., a_k$, which form the input of the program, respectively. The other $N-k$ registers are set to 0. Correspondingly, each of the $NR_i$ variables in our QSIM model has the quantity space $[0, \infty)$. The $NR_i$ variables with nonzero initial values start simulation with qualitative values $<(0, \infty), std>$, whereas the other ones start with $<0, std>$. The quantity space of the variable $U$ is $[0, one]$, where the landmark *one* is equated to 1, as mentioned above. $U$ starts initially at qualitative value $<0, inc>$. The derivative of $U$, $W$, has as quantity space $[0, speed, \infty)$, where *speed* is also equated to 1. It starts at qualitative value $<speed, std>$ and is constant in the whole simulation. The variable $B$ has the quantity space $(-\infty, 0, \infty)$ and starts at $<(0, \infty), dec>$. When started in $OpReg_0$, QSIM will compute a single qualitative behavior segment, which ends with a transition to $OpReg_1$ when $U$ reaches $<one, inc>$ at time-point $t_1$.

---

**Operating Region:** $OpReg_i$

**{Type:** *zero*(*n*)**}**

**Constraint Set:** $d/dt(U, W)$

$NR_n(t) = 0$

All variables except $U$ are constant.

**Possible Transition:**

    **Trigger:** ( $U = <one, inc>$ )

    **New operating region:** $OpReg_{i+1}$

    **Variables inheriting magnitudes or directions:** See Table 5, indexed by the type of $OpReg_{i+1}$

    **Variables with new asserted values:** $U \leftarrow <0, inc>$

---

Table 6: Model Template for Operating Regions Corresponding to *zero*(*n*) Instructions

---

**Operating Region:** $OpReg_i$

**{Type:** *succ*(*n*)**}**

**Constraint Set:** $d/dt(U, W)$

$B(t) + U(t) = NR_n(t)$

All variables except $U$ and $NR_n$ are constant.

**Possible Transition:**

    **Trigger:** ( $U = <one, inc>$ )

    **New operating region:** $OpReg_{i+1}$

    **Variables inheriting magnitudes or directions:** See Table 5, indexed by the type of $OpReg_{i+1}$

    **Variables with new asserted values:** $U \leftarrow <0, inc>$

---

Table 7: Model Template for Operating Regions Corresponding to *succ*(*n*) Instructions

560

Note that there is zero uncertainty about the values of all variables, even the ones with initial magnitude $(0, \infty)$, at the start of the simulation.

Our model is so constrained that a sound and complete qualitative simulator is guaranteed to produce exactly one behavior prediction for any initial state corresponding to a valid URM input. To see this, it is sufficient to observe that, at any step of the simulation, there is sufficient information available to the simulator to compute the exact numerical value of every variable in the model. (This just corresponds to "tracing" the URM program and keeping note of the register contents up to that step.) If the modeled URM halts on the particular input given in the initial state, the QSIM behavior is supposed to be a finite one, ending when the variable $U$ attempts to exceed *one* in $OpReg_{|P|+1}$. If the URM computation does not halt, then the QSIM behavior is supposed to be a single infinite sequence of states, which never visits $OpReg_{|P|+1}$. □

---

**Operating Region:** $OpReg_i$

**{Type:** $jump(m, n, q)$**}**

**Constraint Set:** $d/dt(U, W)$

$NR_m(t) + B(t) = NR_n(t)$

All variables except $U$ are constant.

**Possible Transition:**

    **Trigger:** $( U = \langle one, inc \rangle )$ AND $( B \neq \langle 0, std \rangle )$

    **New operating region:** $OpReg_{i+1}$

    **Variables inheriting magnitudes or directions:** See Table 5, indexed by the type of $OpReg_{i+1}$

    **Variables with new asserted values:** $U \leftarrow \langle 0, inc \rangle$

**Possible Transition:**

    **Trigger:** $( U = \langle one, inc \rangle )$ AND $( B = \langle 0, std \rangle )$

    **New operating region:** $OpReg_q$

    **Variables inheriting magnitudes or directions:** See Table 5, indexed by the type of $OpReg_q$

    **Variables with new asserted values:** $U \leftarrow \langle 0, inc \rangle$

---

Table 8: Model Template for Operating Regions Corresponding to *jump*(*m, n, q*) Instructions

---

**Operating Region:** $OpReg_{|P|+1}$

**{Type:** *End*}

**Constraint Set:** $d/dt(U, W)$

    All variables except $U$ are constant.

---

Table 9: Model of the Operating Region $OpReg_{|P|+1}$, Corresponding to the End of the URM Program

We are now ready to state a new version of the incompleteness theorem.

**Theorem 2:** Even if the qualitative representation is narrowed so that only the *derivative, add, mult,* and *constant* constraints can be used in QDE's, and each variable is forced to start at a finite value with zero uncertainty in the initial state, it is still impossible to build a sound and complete qualitative simulator based on this input-output vocabulary.

**Proof:** Assume that such a sound and complete simulator exists. We now show how to solve the halting problem for URMs using that algorithm as a subroutine.

Construct the corresponding QSIM model as described in Theorem 1 for the URM program P whose halting status on a particular input is supposed to be decided. Now define a new variable *S* with quantity space [0, *one*, ∞), where the landmark *one* is equated to the number 1. *S* starts at the value *<one, std>* in the initial state. Add constraints indicating that *S* is constant to all the operating regions, and specify that the value of *S* is inherited in all possible transitions. Insert the new constraint $S(t) = 0$ in $OpReg_{|P|+1}$. Consider what the simulator is supposed to do when checking the initial state for consistency. Note that we would have an inconsistency if the simulation ever enters $OpReg_{|P|+1}$, since the new constraint that we inserted to that region says that *S* is zero, which would contradict with the inherited value of *one*. So a simulator which is supposed not to make any spurious predictions is expected to reject the initial state at time $t_0$ as inconsistent if the simulation is going to enter $OpReg_{|P|+1}$, in other words, if the URM program under consideration is going to halt. If this sound and complete simulator does not reject the initial state due to inconsistency, but goes on with the simulation, then we can conclude that the program P will not halt. This forms a decision procedure for the halting problem. Since the halting problem is undecidable, we have obtained a contradiction, and conclude that a sound and complete simulator using this representation can not exist. □

It is in fact possible to remove the derivative constraint (which is only used in our proof to ensure that the behavior tree has at most one branch) from the representation as well, and the incompleteness result shown above would still stand:

**Theorem 3:** Even if the qualitative representation is narrowed so that only the *add, mult,* and *constant* constraints can be used in QDE's, and each variable is forced to start at a finite value with zero uncertainty in the initial state, it is still impossible to build a sound and complete qualitative simulator based on this input-output vocabulary.

**Proof:** We will make a minor modification to the proof of Theorem 2. We observe that in the construction of Theorem 1, *U* always starts every operating region at *<0, inc>* and the fact that its derivative is a positive constant forces it to reach the value *<one, inc>* in the next time point. Then the transition to next operating region occurs, and *U* again receives the value *<0, inc>*. What happens if we remove the variable *W* and all *derivative* constraints from the model? In this case, since *U*'s derivative is not fixed, there are three possible states for *U* in the second time point during the simulation of any operating region: *<one, inc>*, *<one, std>*, and *<(0, one), std>*. We fix this problem by inserting another possible region transition specification to all of our regions, except $OpReg_{|P|+1}$. This transition will be triggered when *U* has one of the values *<one, std>*, and *<(0, one), std>*, and its target will be $OpReg_{|P|+1}$. The variable *S* from the proof of Theorem 2, as well as all other variables, are inherited completely during this transition. So all the "unwanted" behaviors which would be created due to the elimination of *U*'s derivative end up in $OpReg_{|P|+1}$, and should therefore be eliminated as spurious in accordance with the argument of the previous proof. Hence, once again, the simulator is supposed to accept the initial state as consistent if and only if P does not halt, meaning that a sound and complete simulation is impossible with this representation as well. □

CAUSES OF INERADICABLE SPURIOUS PREDICTIONS IN QUALITATIVE SIMULATION

Interestingly, one can even restrict the representation so that only nonnegative numbers are supported, and the incompleteness result we proved above still stands:

**Theorem 4:** Even if the qualitative representation is narrowed so that only the *add, mult,* and *constant* constraints can be used in QDE's, each variable is forced to start at a finite value with zero uncertainty in the initial state, and no variable is allowed to have a negative value at any time during the simulation, it is still impossible to build a sound and complete qualitative simulator based on this input-output vocabulary.

**Proof:** In our previous proof, only variable *B* ever has the possibility of having a negative value, and that can occur only in a *jump* region. We replace the definition of the *jump* region template with Table 10, and introduce the new variables *C* and *Y*. We insert constraints that say that these variables are constant to all operating regions. *C* and *Y* start at zero, and are inherited by all transitions, except when the target region is of type *jump*. As can be seen in Table 10, *B* gets the value 1 if and only if the two compared register values are equal. If they are unequal, *B* has a positive value different than 1. In this setup, *B*'s quantity space is defined as [0, *one*, ∞), where *one* is equated to 1, and no variable ever has a negative value during the simulation. *B* now starts simulation with the value <*one*, *dec*> to satisfy the *add* constraint seen in Table 4. The rest of the argument is identical to that of Theorem 3. □

---

**Operating Region:** $OpReg_i$

**{Type:** $jump(m, n, q)$**}**

**Constraint Set:** $NR_m(t) + ONE(t) = C(t)$

$NR_n(t) + ONE(t) = Y(t)$

$B(t) \times C(t) = Y(t)$

All variables except *U* are constant.

**Possible Transition:**

    **Trigger:** ( $U$ = <*one, inc*> ) AND ( $B \neq$ <*one, std*> )

    **New operating region:** $OpReg_{i+1}$

    **Variables inheriting magnitudes or directions:** Depends on the type of $OpReg_{i+1}$, as before

    **Variables with new asserted values:** $U \leftarrow$ <0*, inc*>

**Possible Transition:**

    **Trigger:** ( $U$ = <*one, inc*> ) AND ( $B =$ <*one, std*> )

    **New operating region:** $OpReg_q$

    **Variables inheriting magnitudes or directions:** Depends on the type of $OpReg_q$, as before

    **Variables with new asserted values:** $U \leftarrow$ <0*, inc*>

---

Table 10: Alternative Model Template for Operating Regions Corresponding to *jump*($m, n, q$) Instructions which Avoids Negative Numbers

Alternatively, we can keep negative numbers and remove the *mult* constraint from the representation, if we drop the requirement that each variable starts simulation at a value with zero uncertainty.

**Theorem 5**: Even if the qualitative representation is narrowed so that only the *add* and *constant* constraints can be used in QDE's, it is still impossible to build a sound and complete qualitative simulator based on this input-output vocabulary.

**Proof:** We used the *mult* constraint in the proofs of Theorems 1-3 only for equating variable and landmark values to unambiguous integers. Assume that we delete the *mult* constraints from our model of Theorem 3. The "number" variables of Table 3 are replaced with the setup shown in Table 11. If $R_i$ is supposed to be initialized to the positive integer $a_i$ in P, we equate $NR_i$ to the "$a_i \times unit$" variable in $OpReg_0$ using the method explained in the proof of Theorem 1. Note that we only use *constant* and *add* constraints (and a lot of auxiliary variables) for this purpose.

| CONSTRAINTS | CONCLUSIONS |
|---|---|
| ONEUNIT($t$) = *unit* | |
| ONEUNIT($t$) + ONEUNIT($t$) = TWOUNITS($t$) | TWOUNITS is constant at 2×*unit* |
| ONEUNIT($t$) + TWOUNITS($t$) = THREEUNITS($t$) | THREEUNITS is constant at 3×*unit* |

Table 11: Sample Model Fragment for Equating Variables to Integer Multiples of the Positive Landmark *unit*

The landmarks previously named *one* in other variables' quantity spaces are now equated to *unit*. In this new model, execution of a *succ*($n$) instruction increments $NR_n$'s value by one *unit*. The *jump* instruction compares landmarks whose values equal $u \times unit$ and $v \times unit$ instead of comparing two landmarks whose values equal the natural numbers $u$ and $v$. The *zero* instruction sets the target register to 0, as in the previous construction. So the modeled machine does just what the original URM does, since the multiplication of all values by the coefficient *unit* does not change the flow of the program, and, in particular, whether it halts on its input or not. The rest of the argument is identical to that of the proof of Theorem 3. □

We observe that some of the variables change their qualitative magnitudes and directions discontinuously during operating region transitions in the proofs of the previous theorems. The next theorem proves that maintaining soundness and completeness simultaneously is impossible even if we do not allow any qualitative variable to perform such a change, and force each variable's magnitude and direction to be inherited to the next operating region.

**Theorem 6**: Even if the qualitative representation is narrowed so that only the *derivative*, *add* and *constant* constraints can be used in QDE's, and no variable's magnitude and direction are allowed to perform discontinuous changes during operating region transitions, it is still impossible to build a sound and complete qualitative simulator based on this input-output vocabulary.

**Proof:** Once again, we make some changes to the QSIM models used for simulating the given URM in the previous theorems. As always, we have a QSIM variable $NR_i$ for each of the N registers $R_i$ appearing in the URM program. In addition to that, we define the variables $D_{ij}$ for all $i, j \in \{1, ..., N\}$ such that $i \neq j$. Each of these satisfies the equation $D_{ij} = NR_i - NR_j$ throughout the simulation; that is, we keep track of the differences of all pairs of register values. This can clearly be achieved by inserting several *add* constraints to all the operating regions in our model. These difference variables will enable us to compare two register values in operating regions of type *jump*.

Furthermore, we define auxiliary variables $TR_i$ for all $i \in \{1, ..., N\}$. All the $TR_i$ are initialized to the same values as the corresponding $NR_i$, using the same technique as for the $NR_i$.

For each instruction type in the given URM program, we define *two* operating regions. Our clock variable $U$ will increase in the first of these operating regions to the value <*unit*, *std*>, and decrease in the next one from <*unit*, *std*> to <0, *std*>, performing no discontinuous jump in the program. In order to obtain a variable with such behavior, we make use of the simple harmonic oscillator model given in Table 12, where the variable $X$ (denoting the displacement from the "rest" position of the oscillating "object") oscillates between the values *unit*/2 and −*unit*/2, and the variable $U$ is equated to $X + unit/2$, oscillating between 0 and *unit*. The model template given in Table 12 is added to every operating region. (That table contains some variable names used in the constructions of the previous proofs. All such variables are treated in the previously described manner, unless this proof specifies otherwise.) The following lemma establishes the correctness of this construction.

| CONSTRAINTS | CORRESPONDENCES | MEANING |
|---|---|---|
| $HALFUNIT(t) + HALFUNIT(t) = ONEUNIT(t)$ | $c_1 + c_1 = unit$ | $c_1 = unit/2$ |
| $HALFUNIT(t) + V(t) = E(t)$ | $c_1 + v_1 = 0$ | $v_1 = -unit/2$ |
| $d/dt(X, V)$ | | $\dfrac{dX}{dt} = V$ |
| $d/dt(V, A)$ | | $\dfrac{d^2 X}{dt} = A$ |
| $X(t) + A(t) = Z(t)$ | | $\dfrac{d^2 X}{dt} + X = 0$ |
| $X(t) + HALFUNIT(t) = U(t)$ | | $U = X + unit/2$ |

Table 12: Model Template to Obtain the Desired Behavior for the Variable $U$ as a Clock Oscillating between Qualitative Values <0, *std*> and <*unit*, *std*> (This Template is to be Inserted to All Constructed Operating Regions.)

**Lemma 7:** For any number $r$ which can be represented by a QSIM landmark, a QSIM variable $X$ can be equated to the function $r \sin(t - t_0)$ using only *derivative*, *add* and *constant* constraints.

**Proof of Lemma 7:**
As seen in Table 12, the equation

$$\frac{d^2}{dt} X(t) + X(t) = 0$$

can be expressed using only *derivative*, *add* and *constant* constraints. This equation has a general solution of the form

$$X(t) = c_1 \sin t + c_2 \cos t, \tag{1}$$

and hence its time derivative $V$ has the form

$$V(t) = c_1 \cos t - c_2 \sin t.$$

Assume that $X$ and $V$ are initialized as follows:

$$X(t_0) = 0$$
$$V(t_0) = r.$$

By substituting these values in the equations above, one obtains the equation system

$$0 = c_1 \sin t_0 + c_2 \cos t_0$$
$$r = c_1 \cos t_0 - c_2 \sin t_0,$$

whose solution (Yılmaz, 2005) yields $c_1 = r \cos t_0$ and $c_2 = -r \sin t_0$. Substituting $c_1$ and $c_2$ into Equation (1), we get $X(t) = r \times (\cos t_0 \sin t - \sin t_0 \cos t) = r \sin(t - t_0)$, thereby proving the lemma. □

**Proof of Theorem 6 (continued):**

Therefore, if we equate the landmark $v_1$ of $V$ to $-unit/2$ as shown in Table 12, and initialize $X$ and $V$ to 0 and $v_1$, respectively, we will ensure that

$$X(t) = -\frac{unit}{2} \sin(t - t_0),$$

i.e., that the variable $X$ oscillates between the values $unit/2$ and $-unit/2$, as desired.

To be consistent with Lemma 7, the oscillating variables of Table 12 start simulation with the qualitative values listed in Table 13. All other variables, except $B$, which starts with the value $<(0, \infty), inc>$, are initialized as previously described.

| VARIABLE | QUANTITY SPACE | INITIAL VALUE |
|:---:|:---:|:---:|
| $U$ | [0, *unit*] | $<(0, unit), dec>$ |
| $E$ | $(-\infty, 0, \infty)$ | $<0, std>$ |
| $X$ | $(-\infty, 0, \infty)$ | $<0, dec>$ |
| $V$ | $(-\infty, v_1, 0, \infty)$ | $<v_1, std>$ |
| $A$ | $(-\infty, 0, \infty)$ | $<0, inc>$ |

Table 13: The Quantity Spaces and Initial Values of the Oscillating Variables

We are going to denote the two operating regions corresponding to the $i^{th}$ instruction of the URM program with $OpReg_{i,1}$ and $OpReg_{i,2}$. All variables' qualitative values are inherited in all possible transitions, such that no variable ever undergoes a discontinuous change. Looking carefully at Tables 14-21, which correspond to the initialization, instruction types, and ending of the URM, we see that the simulation flows in a unique branch with the exception of *zero* type operating regions, where there is the possibility that the simulation branches into more than one behavior, and the behaviors which do not correspond to the expected trajectory of the actual URM are directed to $OpReg_{|P|+1,1}$. (Note that transitions to infinite landmarks do not need to be considered, since we assume that all infinite landmarks are specified as unreachable values for all variables in our models.) The registers stay constant when $OpReg_{|P|+1,1}$, which is a single operating region corresponding to the end of the URM program, is reached. The rest of the proof is the same as in Theorem 2. Our contradiction variable $S$ ensures that only the behavior of a non-halting URM leads to a consistent initial state, hence determining the consistency of the initial state is equivalent to deciding the halting problem, leading to a contradiction. □

---

**Operating Region:** $OpReg_0$

**{Type:** *Initialization*}

**Constraint Set:**   All the required "input value representation" constraints (see Table 11)

$B(t) + U(t) = ONEUNIT(t)$ with correspondence "0 + *unit* = *unit*"

*add* constraints linking the $NR_i$ and the $TR_i$ to the relevant "$n \times$unit" variables

*add* constraints defining the $D_{ij}$ variables

the "clock" constraints (Table 12)

All variables except $B$, $U$, $X$, $V$, $E$, and $A$ are constant.

**Possible Transition:**

  **Trigger:** ( $U = <0, std>$ )

  **New operating region:** $OpReg_{1,1}$

  **Variables inheriting qualitative values:** All variables

---

Table 14:  Template for the Single Operating Region Corresponding to the Initialization Stage

## 3.2 Simulation within a Single Operating Region

The incompleteness proofs in subsection 3.1 (as well as that of Say & Akın, 2003) depend on the capability of "turning the constraints on or off" when necessary, which is provided by the operating region transition feature. Would the problem persist if we forsook that feature, and focused on the simulation of qualitative models with a single operating region? We now show that the answer to this question is affirmative.

### 3.2.1 HILBERT'S TENTH PROBLEM

As the name suggests, Hilbert's Tenth Problem is the tenth of 23 problems which were announced in 1900 by the famous mathematician David Hilbert as a challenge to the mathematicians of the 20[th] century. It asks for an algorithm for deciding whether a given multivariate polynomial with integer coefficients has integer solutions. It has been proven that no such algorithm exists (Matiyasevich, 1993). This fact was used by Say and Akın (2003) in their original proof of the existence of ineradicable spurious predictions in the outputs of all qualitative simulators employing the operating region transition feature and a larger set of constraint types than those we deal with in this paper.

In the proof to be presented shortly, we use the undecidability of a slightly modified variant of the setup described by Hilbert: We assume a guarantee that none of the variables in the given polynomial are zero in the solution whose existence is in question. It is clear that this modified problem is unsolvable as well, by the following argument: Assume that we do have an algorithm A which takes a multivariate polynomial with integer coefficients as input, and announces whether a solution where all the variables have nonzero integer values exists or not in finite time. We can use A as a subroutine in the construction of the algorithm sought in Hilbert's original problem as follows: We systematically produce $2^n$ polynomials from the input polynomial with $n$ variables, such that each of these new polynomials corresponds to a different subset of the variables of the original polynomial replaced with zero. We then run A on each of these new polynomials. It is easy to see that A will find that one or more of these polynomials have nonzero integer solutions if and only if the original polynomial has integer solutions.

---

**Operating Region:** $OpReg_{i,1}$

**{Type:** $zero(n)$**}**

**Constraint Set:**  *add* constraints defining the $D_{ij}$ variables

  the "clock" constraints (Table 12)

  All variables except $U$, $X$, $V$, $E$, $A$, $NR_n$, and $D_{ij}$ with $n \in \{i,j\}$ are constant.

**Possible Transition:**

  **Trigger:** ( $U$ = <*unit*, *std*> ) AND ( $NR_n$ = <0, *std*> )

  **New operating region:** $OpReg_{i,2}$

  **Variables inheriting qualitative values:** All variables

**Possible Transition:**

  **Trigger:** (( $U$ = <*unit*, *std*> ) AND ( $NR_n \neq$ <0, *std*> )) OR ( $NR_n$ = <(0, ∞), *std*> ) OR ( $NR_n$ = <(-∞, 0), *std*> )

  **New operating region:** $OpReg_{|P|+1, 1}$

  **Variables inheriting qualitative values:** All variables

---

Table 15: Template for the First Operating Region Corresponding to *zero*(*n*) Instructions

### 3.2.2 SOUND AND COMPLETE QSIM WITHIN A SINGLE OPERATING REGION SOLVES HILBERT'S TENTH PROBLEM

**Theorem 8**: Even if the qualitative representation is narrowed so that only the *derivative*, *add*, *mult* and *constant* constraints can be used in QDE's, and the simulation proceeds only in a single operating region, it is still impossible to build a sound and complete qualitative simulator based on this input-output vocabulary.

We are going to start our proof with some preliminary lemmata, the first of which is reminiscent of Lemma 7 from the previous subsection:

**Lemma 9**: For any real constant equated to the QSIM variable $X_i$, a QSIM variable $Y_i$ can be equated to the function $\sin(X_i \times (t - t_0))$ using only *derivative*, *add*, *mult*, and *constant* constraints.

**Proof:** The case for $X_i = 0$ is trivial, and can be handled with a single *constant* constraint. For the remaining case, we will consider the following equation set:

---

**Operating Region:** $OpReg_{i,2}$

**{Type:** $zero(n)$**}**

**Constraint Set:** *add* constraints defining the $D_{ij}$ variables

the "clock" constraints (Table 12)

All variables except $U$, $X$, $V$, $E$, $A$, and $TR_n$ are constant.

**Possible Transition:**

    **Trigger:** ( $U$ = <0, *std*> ) AND ( $TR_n$ = <0, *std*> )

    **New operating region:** $OpReg_{i+1,1}$

    **Variables inheriting qualitative values:** All variables

**Possible Transition:**

    **Trigger:** (( $U$ = <0, *std*> ) AND ( $TR_n \neq$ <0, *std*> )) OR ( $TR_n$ = <(0, ∞), *std*> ) OR ( $TR_n$ = <(-∞, 0), *std*> )

    **New operating region:** $OpReg_{|P|+1,\ 1}$

    **Variables inheriting qualitative values:** All variables

---

Table 16: Template for the Second Operating Region Corresponding to *zero*(*n*) Instructions

---

**Operating Region:** $OpReg_{i,1}$

**{Type:** $succ(n)$**}**

**Constraint Set:** $TR_n(t) + U(t) = NR_n(t)$

*add* constraints defining the $D_{ij}$ variables

the "clock" constraints (Table 12)

All variables except $U$, $X$, $V$, $E$, $A$, $NR_n$, and $D_{ij}$ with $n \in \{i,j\}$ are constant.

**Possible Transition:**

    **Trigger:** ( $U$ = <*unit*, *std*> )

    **New operating region:** $OpReg_{i,2}$

    **Variables inheriting qualitative values:** All variables

---

Table 17: Template for the First Operating Region Corresponding to *succ*(*n*) Instructions

$$\frac{d^2}{dt}Y_i(t) + W_i \times Y_i(t) = 0 \qquad (2)$$

$$W_i = X_i^2, \text{ so that } \sqrt{W_i} = \begin{cases} X_i, & X_i > 0 \\ -X_i, & X_i < 0. \end{cases} \qquad (3)$$

with the initial values

$$Y_i(t_0) = 0$$

---

**Operating Region:** $OpReg_{i,2}$
**{Type:** $succ(n)$**}**
**Constraint Set:** $TR_n(t) + U(t) = NR_n(t)$
                 add constraints defining the $D_{ij}$ variables
                 the "clock" constraints (Table 12)
                 All variables except $U$, $X$, $V$, $E$, $A$, and $TR_n$ are constant.
**Possible Transition:**
    **Trigger:** ( $U = <0, std>$ )
    **New operating region:** $OpReg_{i+1,1}$
    **Variables inheriting qualitative values:** All variables

Table 18: Template for the Second Operating Region Corresponding to *succ*(*n*) Instructions

---

**Operating Region:** $OpReg_{i,1}$
**{Type:** $jump(m, n, q)$**}**
**Constraint Set:** add constraints defining the $D_{ij}$ variables
                 the "clock" constraints (Table 12)
                 All variables except $U$, $X$, $V$, $E$, and $A$ are constant.
**Possible Transition:**
    **Trigger:** ( $U = <unit, std>$ )
    **New operating region:** $OpReg_{i,2}$
    **Variables inheriting qualitative values:** All variables

Table 19: Template for the First Operating Region Corresponding to *jump*(*m*, *n*, *q*) Instructions

$$V_i(t_0) = X_i,$$

where $V_i(t)$ is the time derivative of $Y_i(t)$.

The general solution of Equation (2) is:

$$Y_i(t) = c_1 \sin\left(\sqrt{W_i} \times t\right) + c_2 \cos\left(\sqrt{W_i} \times t\right).$$

Substituting the $\sqrt{W_i}$ from Equation (3) and the initial values in the equations for $Y_i$ and $V_i$, and solving these equation systems results in the following (Yılmaz, 2005):

        For $X_i > 0$,   $c_1 = \cos(X_i \times t_0)$ and $c_2 = -\sin(X_i \times t_0)$.

        For $X_i < 0$,   $c_1 = -\cos(X_i \times t_0)$ and $c_2 = -\sin(X_i \times t_0)$.

When we substitute these in the formula for $Y_i(t)$, we obtain:

$$Y_i(t) = \cos(X_i \times t_0)\sin(X_i \times t) - \sin(X_i \times t_0)\cos(X_i \times t) = \sin(X_i \times (t - t_0)), \quad\quad X_i > 0,$$

---

**Operating Region:** $OpReg_{i,2}$

**{Type:** $jump(m, n, q)$**}**

**Constraint Set:**  *add* constraints defining the $D_{ij}$ variables

         the "clock" constraints (Table 12)

         All variables except $U$, $X$, $V$, $E$, and $A$ are constant.

**Possible Transition:**

    **Trigger:** ( $D_{mn} = <0, std>$ ) AND ( $U = <0, std>$ )

    **New operating region:** $OpReg_{q,1}$

    **Variables inheriting qualitative values:** All variables

**Possible Transition:**

    **Trigger:** ( $D_{mn} \neq <0, std>$ ) AND ( $U = <0, std>$ )

    **New operating region:** $OpReg_{i+1,1}$

    **Variables inheriting qualitative values:** All variables

---

Table 20: Template for the Second Operating Region Corresponding to $jump(m, n, q)$ Instructions

---

**Operating Region:** $OpReg_{|P|+1, 1}$

**{Type:** *End*}

**Constraint Set:**  $S(t) = 0$

         All variables except $U$, $X$, $V$, $E$, and $A$ are constant.

---

Table 21: Model of the Operating Region Corresponding to the End of the URM Program

$$Y_i(t) = -\cos(X_i \times t_0)\sin((-X_i) \times t) - \sin(X_i \times t_0)\cos((-X_i) \times t) = \sin(X_i \times (t - t_0)), \qquad X_i < 0.$$

Hence, we have $Y_i(t) = \sin(X_i \times (t - t_0))$ for all $X_i \neq 0$.

Table 22 shows that Equations (2) and (3) and the initial values given above are representable using the *derivative*, *add*, *mult*, and *constant* constraints. Note that $X_i$ has to be kept constant and initialized to either $(0, \infty)$ or $(-\infty, 0)$, depending on the intended sign for that number, and both $Y_i$ and $C_i$ must start at zero, to be consistent with the construction above. □

**Lemma 10:** Starting at $t_0$, the function $Y = \sin(t-t_0)$ reaches the value 0 for the first time at time point $t_E = t_0 + \pi$. Moreover the function $Y_i = \sin(X_i \times (t-t_0))$ reaches 0 at the same time point $t_E$ if and only if $X_i$ is an integer.

**Proof:** The equation $\sin(t-t_0) = 0$ implies $t-t_0 = n\pi$, $n \in \mathbb{Z}$, and since we are interested in the first time point after $t_0$ where it becomes 0, we get $t_E = t_0 + \pi$. For the "only if" part of the second statement, assume that the function $Y_i = \sin(X_i \times (t-t_0))$ reaches 0 at $t_E = t_0 + \pi$. Then $\sin(X_i \times (t_0 + \pi - t_0)) = \sin(X_i \times \pi) = 0$ implies that $X_i$ is an integer. For the "if" part, we use the knowledge that $X_i$ is an integer to conclude that $Y_i(t_E) = \sin(X_i \times (t_0 + \pi - t_0)) = \sin(X_i \times \pi) = 0$. □

YILMAZ & SAY572572YILMAZ & SAY

| CONSTRAINTS | MEANING |
|---|---|
| $Z = 0$ | |
| $X_i(t) + C_i(t) = V_i(t)$ | $V_i(t_0) = X_i$ |
| $d/dt(Y_i, V_i)$ | $\dfrac{dY_i}{dt} = V_i(t)$ |
| $d/dt(V_i, A_i)$ | $\dfrac{d^2 Y_i}{dt} = A_i(t)$ |
| $X_i(t) \times X_i(t) = W_i(t)$ | $W_i = X_i^2$ |
| $W_i(t) \times Y_i(t) = L_i(t)$ | $L_i(t) = W_i \times Y_i(t)$ |
| $A_i(t) + L_i(t) = Z(t)$ | $\dfrac{d^2 Y_i}{dt} + W_i \times Y_i(t) = 0$ |

Table 22: Model Fragment Used to Obtain the Relationship $Y_i = \sin(X_i \times (t - t_0))$

**Proof of Theorem 8**: As already mentioned, the proof relies on a contradiction, namely that a sound and complete simulator, if it existed, could be used to construct an algorithm for solving Hilbert's Tenth Problem, as follows:

Assume that we are given a polynomial $P(x_1, x_2, x_3, ..., x_n)$ with integer coefficients. We start by constructing a QSIM model fragment that "says" that $P(x_1, x_2, x_3, ..., x_n) = 0$: We have already seen in Section 3.1 how to equate any desired integer to a QSIM variable. Represent all integers appearing as coefficients in the polynomial in that manner. Introduce a QSIM variable $X_i$ for each of the $x_i$, declare all these $X_i$ to be constant, and use *add* and *mult* constraints to equate the sum of products that is $P(X_1, X_2, X_3, ..., X_n)$ to a QSIM variable $P$, which will be initialized to 0. Note that this is tantamount to saying that the present values of the $X_i$ form a solution for the polynomial. All this can clearly be done in a single operating region, with constraints of the types *mult*, *add* and *constant*.

We then extend this model with the necessary constraints and auxiliary variables to equate a new variable $Y$ to the function $\sin(t-t_0)$. (Either Lemma 7 or Lemma 9 can be used for this purpose.) We specify $Y$'s quantity space as $[0, \infty)$, so that the simulation is guaranteed to finish at $t = t_E = t_0 + \pi$. For each $X_i$, we define associated auxiliary variables $C_i$, $L_i$, $W_i$, $V_i$, $A_i$ and $Y_i$, and add the template in Table 22 to our model to express the relationship $Y_i = \sin(X_i \times (t-t_0))$. We also equate a variable $YS$ with the sum of the squares of the $Y_i$, i.e. $YS = \sum_{i=1}^{n} Y_i^2$. Note that if $YS = 0$, then all the $Y_i$ are 0.

Finally, we need make sure that the only consistent behaviors are the ones in which the $X_i$ are integers (that is, relying on Lemma 10, the behaviors in which the variable $YS$ becomes 0 at $t_E$). To serve this aim, we add the constraint $F(t) \times Y(t) = YS(t)$ to our model.

We will simulate this model $2^n$ times, each run corresponding to a different way of initializing the $X_i$ to magnitudes selected from the set $\{(0, \infty), (-\infty, 0)\}$. A sound and complete simulator

572572

would accept all and only the initial states with those $X_i$ whose values do not cause any inconsistency with our model. But those $X_i$ correspond exactly to the integer solutions of the given polynomial, by the following reasoning about the variable $F$:

Note that $F$ is defined to be $YS/Y$ by the $F(t) \times Y(t) = YS(t)$ constraint. We know that $Y$, $Y_i$, and hence $YS$ are all initially 0, meaning that one has to use l'Hôpital's rule to find out the initial value of $F$. This is important, since if $F$'s initial magnitude or derivative are infinite, QSIM is not even supposed to consider successors for the initial state. (We declare the infinite landmarks as unreachable values for all variables, as mentioned earlier. Even if $F$'s magnitude is finite and just its derivative is infinite, simulation is not supposed to continue, because the continuity requirement would be violated.) Fortunately,

$$F(t) = \frac{\sum_{i=1}^{n} \sin^2(X_i \times (t - t_0))}{\sin(t - t_0)}$$

does have finite magnitude and derivative at $t_0$: At $t = t_0$, we use l'Hôpital's rule to find

$$F(t_0) = \frac{0}{0} = \frac{\sum_{i=1}^{n} 2X_i \sin(X_i \times (t_0 - t_0)) \cos(X_i \times (t_0 - t_0))}{\cos(t_0 - t_0)} = 0.$$

As for $F$'s qualitative direction;

$$\frac{dF}{dt}(t) = \sum_{i=1}^{n} \left( \frac{2X_i \sin(X_i \times (t - t_0)) \cos(X_i \times (t - t_0))}{\sin(t - t_0)} - \frac{\sin^2(X_i \times (t - t_0)) \cos(t - t_0)}{\sin^2(t - t_0)} \right),$$

and it turns out, after several applications of l'Hôpital's rule, that

$$\frac{dF}{dt}(t_0) = \sum_{i=1}^{n} X_i^2,$$

which is clearly a finite positive number, and $F$'s initial qualitative value is therefore <0, *inc*>.

Obviously, $F = YS/Y$ is guaranteed to be finite until $t_E$, when $Y$ reaches 0. If the variable $YS$ is nonzero (implying that at least one of the $Y_i$ is nonzero, and by Lemma 10 that the corresponding $X_i$ is not an integer) at $t_E$, $F(t_E)$ has to equal $\infty$, which is impossible since infinity was declared to be unreachable, so such states would be eliminated as spurious. If, on the other hand, $YS(t_E) = 0$, then, we see by l'Hôpital's rule, and the knowledge that all the $X_i$ *are* integers, that

$$F(t_E) = \frac{0}{0} = \frac{\sum_{i=1}^{n} 2X_i \sin(X_i \times (t_E - t_0)) \cos(X_i \times (t_E - t_0))}{\cos(t_E - t_0)} = \frac{\sum_{i=1}^{n} 2X_i \sin(X_i \times \pi) \cos(X_i \times \pi)}{\cos(\pi)} = 0,$$

and behaviors ending with such states are supposed to be included in the simulation output. So if our supposedly complete and sound simulator rejects the initial states of our model due to inconsistency in all the $2^n$ runs, we reason that all the behavior predictions considered by the simulator ended with $F(t_E) = \infty$, and this inconsistency propagated back to the initial state and led to its rejection in all cases. We conclude that the polynomial has no integer solutions. On the other hand, if even one of the simulations prints out the initial state and goes on with its successors, we conclude that a solution exists. This forms the decision procedure required in Hilbert's Tenth Problem, leading to a contradiction. Therefore, sound and complete simulation is impossible even if one restricts oneself to a single operating region and the limited constraint vocabulary mentioned in the statement of the theorem. □

## 4. Conclusion

In this paper, we considered several alternative subsets of the qualitative representation, and showed that the ineradicable spurious prediction problem persists even when only the *add* and *constant* constraints are allowed. If one allows the *mult* constraint as well, then any resulting qualitative simulator is inherently incomplete even when the representation of negative numbers is forbidden and every variable is forced to be specified with zero uncertainty (i.e. as a single unambiguous real number) in the initial state. Our final proof shows that even the ability of handling models with multiple operating regions can be removed from the representation, and the incompleteness problem would still persist, provided the *add*, *constant*, *derivative*, and *mult* constraints are allowed in the vocabulary. Note that none of these vocabularies include the monotonic function constraint, which is the only relation type "native" to the qualitative representation.

Although the results in this paper are demonstrated using the QSIM representation for input and output, they are valid for all qualitative simulators whose input and output vocabularies are as expressive as the specified subsets of those of QSIM. (Also note that our proofs apply automatically to semi-quantitative simulators, whose representations are an extension of that of pure QSIM.) We believe that these results are important in the sense that they provide deeper insight to the causes of spurious predictions, and they can be helpful for researchers aiming to construct provably sound and complete simulators using weaker representations.

Finally, we wish to stress that our findings here do not amount to as bad a piece of news about the usefulness of qualitative simulators in the practical domains that they are usually utilized as it may seem to the uninitiated eye. When one's model is specified at the level of precision that is involved in the models in this paper, one does not employ a *qualitative* reasoner anyway. What really annoys the users of qualitative simulators is the occasional prediction of *eradicable* spurious behaviors, and the strengthening of the algorithms with additional filters of increasing mathematical sophistication to get rid of more of these continues to be an important line of research.